%% file: main.tex
\documentclass[preprint,nopreprintline,10pt,authoryear]{elsarticle}
\input{preamble}
\begin{document}
\begin{frontmatter}
\title{\makebox[\textwidth][c]{Physiological Trajectories and Observation Patterns in ICU Mortality Prediction:}\\Transportability Under Changing Measurement Conditions}
\author{Benjamin T. Pike}
\begin{abstract}
\input{sections/abstract}
\end{abstract}
\end{frontmatter}
\input{sections/introduction}
\input{sections/methods}
\input{sections/results}

\input{sections/discussion}
\input{sections/conclusion}
\input{sections/declarations}
\FloatBarrier
\bibliographystyle{elsarticle-num-names}
\bibliography{references}
\end{document}

%% file: preamble.tex
\usepackage[british]{babel}
\usepackage[a4paper,margin=2cm]{geometry}
\usepackage{lineno}
\usepackage{hyperref}
\hypersetup{
    colorlinks=true,
    linkcolor=blue,
    citecolor=blue,
    urlcolor=blue,
    pdftitle={Physiological Trajectories and Observation Patterns in ICU Mortality Prediction: Transportability Under Changing Measurement Conditions}
}
\usepackage{amssymb,amsmath}
\usepackage{svg}
\usepackage{graphicx}
\usepackage{tabularx,booktabs,array,ragged2e,threeparttable}
\usepackage{caption,subcaption}
\usepackage{float}
\usepackage{xargs}
\usepackage{multirow}
\usepackage[prependcaption]{todonotes}
\newcolumntype{L}[1]{>{\RaggedRight\arraybackslash}p{#1}}
\newcommand{\tableformat}[1][6pt]{\centering\small\setlength{\tabcolsep}{#1}}
\newcommand{\cirange}[2]{#1 to #2}
\setcitestyle{numbers,square,comma}

\usepackage{placeins}
\usepackage{xurl}
\graphicspath{{figures/images/}}

%% file: sections/abstract.tex
\textbf{Objective} To examine how changes in clinical measurement practices affect the transportability of predictive information in longitudinal ICU records, distinguishing physiological trajectories from patterns of observation.

\textbf{Methods} Using measurements recorded within the first 48 hours of ICU admission, in-hospital mortality was predicted in MIMIC-III, MIMIC-IV, and eICU. Nested representations combined the latest recorded physiological state (R1), summaries of physiological trajectories (R2), and features capturing the timing and frequency of observations (R3), allowing their incremental predictive value to be compared across settings. Models trained on 7,762 MetaVision patients were evaluated using 21,682 MIMIC-IV patients and 43,878 eICU patients from 204 hospitals. Restriction experiments examined how the temporal distribution and availability of measurements affected prediction, holding observation counts constant for timing comparisons and assessing whether adaptation could recover performance under reduced availability.

\textbf{Results} Longitudinal physiological information improved prediction beyond latest recorded values across all evaluation settings, including following transfer to eICU (R2--R1 log-loss difference $-0.01446$, 95\% CI $-0.01701$ to $-0.01202$). Observation patterns improved prediction following transfer from MetaVision to MIMIC-IV (R3--R2 $-0.00564$) and in locally developed eICU models (R3--R2 $-0.00402$, 95\% CI $-0.00517$ to $-0.00283$), but provided no clear benefit following transfer from MetaVision to eICU ($+0.00106$, 95\% CI $-0.00087$ to $+0.00355$). With observation counts held constant, retaining measurements spread throughout the 48-hour window reduced log loss relative to random retention, particularly for R3 (0.04324 versus 0.00721 for R2). R3 was also more sensitive to the loss of recent measurements. Although reduced observation availability eliminated R3's advantage over R2, retraining under the same conditions restored it.

\textbf{Conclusions} Physiological trajectories transferred consistently, while observation patterns provided less reliable benefits and were more sensitive to measurement timing and availability. Their transferability may depend on whether training captures sufficiently diverse measurement practices to generalise across settings. Clinical prediction models should therefore be evaluated across measurement conditions as well as patient populations.

%% file: sections/introduction.tex
\section{Introduction}

Modern intensive care produces extensive longitudinal records, with repeated physiological measurements detailing not only a patient’s current state but also how their condition evolves over time. Latest measurements of heart rate, blood pressure, creatinine, or lactate offer only a snapshot, without revealing whether an abnormality is persistent, improving, or worsening. Models tasked with predicting ICU mortality often make use of repeated observations by incorporating the longitudinal information into their feature sets \citep{silva2012,harutyunyan2019,sun2025}. Large ICU datasets, such as MIMIC and eICU, facilitate development and evaluation of such models across different time periods, recording systems, and hospitals \citep{johnson2016,johnson2023,pollard2018}.

Beyond the values themselves, longitudinal records also capture the patterns by which patients are observed. Two patients may exhibit broadly similar values but differ substantially in the timing, frequency, and regularity of these measurements. Such differences may reflect clinical concern, treatment protocols, or local monitoring practices. Consequently, observation patterns themselves may carry predictive value. This phenomenon has been termed informative presence or informative observation \citep{agniel2018,sisk2021}, with models such as GRU-D explicitly incorporating this information alongside the measurement values \citep{che2018}.

However, the extent to which observation patterns generalise across settings remains unclear. While the physiological trajectory of a patient may retain broadly similar meaning, the timing and frequency of measurements are in part determined by the local environment. Associations learned between these patterns and outcomes may therefore be valuable when models are evaluated in their development setting but ineffective, or even detrimental, elsewhere. This concern, where prediction is dependent on the stability of the measurement process, has been previously described by \citet{groenwold2020}. More recently, features that capture observation patterns have been found to weaken the transportability of sepsis mortality models \citep{yamamoto2026} and add little value beyond trajectories in ICU acute kidney injury prediction \citep{zou2026}.

Differences in model performance across settings can be difficult to interpret because patient populations, available variables, and measurement practices may change simultaneously \citep{riley2016,pajouheshnia2019}. The volume of observations in a patient's record also tells us little about when they were made. Two records may contain the same number of measurements but differ substantially in when they were recorded. Measurements may be concentrated at different points across the preceding history, with the timing and degree of clustering varying across records. It is unclear whether such variation influences prediction independently of observation count, and whether models that incorporate observation patterns are particularly sensitive to it.

At 48 hours after ICU admission, this study examines whether longitudinal physiological summaries improve mortality prediction beyond the latest recorded values. It then assesses whether observation patterns provide additional predictive information and whether this information generalises across settings. The study also investigates whether the timing of measurements affects performance when the number of observations is held constant. Further local development and explanatory analyses explore possible reasons for differences in performance, while adaptation analyses examine whether performance can be recovered when observation availability changes.

%% file: sections/methods.tex
\section{Methods}

\subsection{Prediction task and study overview}

The study examined the prediction of in-hospital mortality using measurements taken within the first 48 hours of an ICU stay. Patients whose stay was shorter than 48 hours were excluded. Data were sourced from the single-institution MIMIC-III (v1.4) and MIMIC-IV (v3.1) cohorts, and from the multicentre eICU (v2.0) database. MIMIC-III contains data derived from two recording systems, CareVue and MetaVision, and overlaps with MIMIC-IV. Here, CareVue and MetaVision refer to the corresponding MIMIC-III stays, whilst MIMIC-IV refers only to stays recorded after the MIMIC-III period.

Five nested representations were created to assess how different aspects of the patient record contributed to prediction. Starting with a minimal baseline, these progressively incorporated the latest recorded physiological state, longitudinal summaries of trajectory, severity, and variability, information relating to the timing and frequency of observations, and a wider set of physiological and observation characteristics. Using the MIMIC datasets, generalisation within a single institution across recording systems and periods was evaluated first. Models were then trained on all eligible MetaVision stays and evaluated on MIMIC-IV and eICU. Finally, to examine the impact of measurement quantity and timing on prediction, restriction experiments systematically removed observations from the patient record and rebuilt the representations.

\subsection{Populations and evaluation settings}

All cohorts were restricted to adults aged 18 years or older. The CareVue cohort included 12,518 eligible patients, whose records were used to train models evaluated on all 7,762 eligible MetaVision patients. To avoid overlap with the MIMIC-III recording period, MIMIC-IV was restricted to ICU stays with an earliest possible year of 2013 or later. Models trained on the MetaVision patients were then evaluated on all 21,682 eligible MIMIC-IV patients. Further details of cohort selection are provided in the Supplementary Material.

The eICU cohort additionally required a recorded hospital outcome. One hospital admission was selected per patient, followed by the earliest ICU stay within that admission. Applying the eligibility criteria yielded 54,787 patients from 205 hospitals, of which 10,909 were reserved for adaptation analyses, while the remaining 43,878 served as the external test set. The partitions were disjoint at the patient level, but not at the hospital level. Cohort characteristics are presented in Table~\ref{tab:cohorts}.

\input{tables/cohorts}

Within each source cohort, the same five outcome-stratified folds were used across all representations for internal evaluation and out-of-fold calibration. Models developed locally in eICU were evaluated using hospital-grouped cross-validation, repeated over 20 allocations, across which performance and paired differences were averaged equally.

The representation hierarchy was specified before examining performance in eICU. MetaVision models were evaluated using the MIMIC-IV and eICU cohorts without additional tuning or refitting. Further analyses, including model adaptation and harmonisation sensitivity, were conducted post hoc.

\subsection{Model inputs and representations}

The five nested representations, R0--R4, were developed as summarised in Table~\ref{tab:tiers}. R0 serves as a minimal baseline, while R1 represents the patient's state at 48 hours, predominantly using the latest recorded measurements and other scalar predictors. R2 extends this snapshot with longitudinal summaries describing physiological severity, trends, and variability, whilst R3 adds features describing the observation process, including the frequency and timing of measurements. R4 incorporates a broader range of physiological and observation summaries. Each representation retains all features from the preceding tier.

\input{tables/tiers}

Representations were compared to determine whether longitudinal information improved prediction beyond the latest recorded physiological state (R2--R1) and the potential additional value of observation patterns (R3--R2). Each representation changes the size and structure of the model input, which affect feature interactions and regularisation. Consequently, these comparisons consider the contribution of each representation as a whole rather than the isolated effects of additional features.

For external transfer, variables were harmonised between MetaVision and eICU following manual review and mapping of the corresponding database tables, variable definitions, and measurement units. Ventilation status and all dependent features were excluded from the shared representations because a comparable definition of a ventilation episode could not be established in eICU. Likewise, Glasgow Coma Scale (GCS) measurements were excluded from all representations due to differences in component availability, coding, and handling of responses that could not be assessed. Complete feature definitions, mappings, missingness rules, and exclusions are available in the Supplementary Material.

\subsection{Observation restriction and timing analyses}

To characterise the sensitivity of model performance to observation availability, measurements were removed at random from the patient record. Measurements were thinned to 50\%, and subsequently 25\%, of their original counts, followed by recalculation of all derived features. Static values recorded on admission to the ICU remained unchanged. 

Sensitivity to timing was estimated by holding the total count of retained observations constant per patient. Approximately 25\% of measurements were selected either at random, preferentially from the most recent observations, or distributed throughout the 48-hour window. The relative importance of recent measurements was determined by ablating the final 12 hours of values and contrasting performance against random removal of the same number of measurements.

Collectively, these experiments sought to distinguish between the effects of observation quantity and timing. Randomised restrictions were repeated across ten trials, with the same iterations used for paired comparisons. Following each restriction, model inputs were reconstructed and evaluated using the existing models without refitting.

\subsection{Model fitting, adaptation, and supporting analyses}

CareVue and MetaVision models were developed using five outcome-stratified folds, within which preprocessing and model fitting were performed separately. Uncalibrated out-of-fold predictions were used to evaluate internal performance and, together with observed outcomes, to estimate intercept-only and Platt calibration adjustments. Models were then trained on the full eligible source cohorts for external evaluation. The effect of each change in representation was averaged over 20 training runs. All primary comparisons are based on uncalibrated predictions, while calibrated results for MetaVision models are reported separately. To assess whether performance depended on the modelling approach, R1--R3 models were also trained on the MetaVision cohort using histogram gradient boosting, depth-two XGBoost, random forest, and penalised logistic regression.

Three strategies were investigated to adapt models to reduced observation availability. First, eICU models were retrained on records with observation availability restricted to match that of the evaluation data. Second, existing predictions were recalibrated using intercept-only or slope-and-intercept adjustments estimated from eICU calibration patients, using either complete records or records subjected to the same restriction. Third, MetaVision R2 and R3 predictions were combined using eICU calibration data, allowing the relative contribution of the latter to vary with availability.

Post-hoc analyses examined reverse transfer from eICU to MetaVision and MIMIC-IV. Shared R1--R3 models were trained on the full eICU population and on samples matched to MetaVision size and mortality, then evaluated unchanged on both targets. Models were also trained on the combined MetaVision and MIMIC-IV cohorts and evaluated using eICU. Full experimental details are available in the Supplementary Material.

\subsection{Evaluation}

Mean log loss serves as the primary evaluation metric. AUROC, Brier score, and calibration are reported as secondary measures. In paired evaluations, negative log-loss differences favour the named candidate.

Confidence intervals were calculated using paired bootstrap resampling. For eICU analyses, whole hospitals were resampled to account for clustering within institutions. Performance was reported using both patient-level and equal-hospital weighting. Fixed out-of-fold predictions were used for internal source comparisons, with patients resampled to estimate confidence intervals for paired differences. For local eICU analyses, estimates were averaged equally across hospital allocations, retaining each resampled hospital across all allocations in which it appeared.

Stability of the external results to variation in MetaVision model fitting was assessed using 30 bootstrap refits of the source cohort, with each evaluated on the eICU test set. To simultaneously account for variance in source training and target hospital composition, a joint sensitivity analysis was performed, which combined these source refits with 10,000 hospital-level resamples of the eICU data. For supporting analyses, $p$-values were adjusted using Holm's method within predefined comparison families.

%% file: tables/cohorts.tex
\begin{table}[H]
\tableformat[10pt]
\caption{Eligible MIMIC cohorts and the eICU evaluation population. Age is reported as median (interquartile range) in years. CareVue, MetaVision, and MIMIC-IV relate to the same institution. Seven eICU test records have sex outside the female/male categories. The separate 10,909-patient adaptation population is not included.}
\label{tab:cohorts}
\begin{tabular}{lrrrrr}
\toprule
Cohort & Patients & Deaths (\%) & Age & Female & Hospitals \\
\midrule
CareVue & 12,518 & 1,733 (13.8) & 67 (53--78) & 5,488 & 1 \\
MetaVision & 7,762 & 1,005 (12.9) & 66 (54--79) & 3,485 & 1 \\
MIMIC-IV & 21,682 & 2,978 (13.7) & 67 (56--77) & 9,387 & 1 \\
eICU & 43,878 & 5,393 (12.3) & 66 (55--77) & 19,909 & 204 \\
\bottomrule
\end{tabular}
\end{table}

%% file: tables/tiers.tex
\begin{table}[H]
\tableformat
\caption{Nested information representations. Feature counts are reported before categorical variables are expanded during encoding. Shared counts indicate the subset retained after harmonisation between MetaVision and eICU.}
\label{tab:tiers}
\begin{tabularx}{\linewidth}{@{}lL{3cm}ccX@{}}
\toprule
Tier & Interpretation & MIMIC & Shared & Information added \\
\midrule
R0 & Baseline & 10 & 10 & Static inputs and latest values of six physiological variables. \\
R1 & Snapshot & 42 & 41 & Additional scalar values, urine output, and ventilation indicators. \\
R2 & Trajectory & 326 & 325 & Physiological severity, trend, and variability summaries. \\
R3 & Observation patterns & 510 & 500 & Measurement frequency, timing, and observation-pattern summaries. \\
R4 & Rich summary & 915 & 893 & Additional physiological and observation summaries. \\
\bottomrule
\end{tabularx}
\end{table}

%% file: sections/results.tex
\section{Results}

\subsection{Physiological trajectories improved prediction and transferred more consistently}

Across the MIMIC cohorts, longitudinal physiological information (R2) improved prediction beyond latest value snapshots (R1). In MIMIC-IV, AUROC increased from 0.8486 to 0.8704, while log loss decreased by 0.01355. Observation patterns (R3) increased AUROC further to 0.8759, with little additional benefit provided by the richest feature set (R4). The CareVue-to-MetaVision transfer and internal source evaluations demonstrated a similar pattern.

The advantage of longitudinal information over snapshots persisted following external transfer of the MetaVision models to the 43,878-patient eICU test set. Log loss dropped from 0.30872 (R1) to 0.29426 (R2), with corresponding AUROCs of 0.79049 and 0.81920 (Table~\ref{tab:external}). These results were consistent across weighting schemes. Among the 79 eICU hospitals with at least 100 patients, 20 deaths, and 20 survivors, R2 achieved lower log loss in 91.1\%, compared with 76.0\% of all hospitals.

These findings were consistent across source refits, with R2 outperforming R1 in all 30 bootstrap refits of the MetaVision cohort. Histogram gradient boosting, depth-two XGBoost, and random forest reproduced the direction of the external transfer result under both weighting schemes. Penalised logistic regression behaved differently, with performance deteriorating as the representations became more complex.

\input{tables/external_performance}

\subsection{Observation patterns were predictive but transferred less consistently}

Although observation patterns improved prediction beyond longitudinal information in MIMIC-IV, this benefit did not cleanly transfer to eICU. In MIMIC-IV, R3 reduced log loss by 0.00564 relative to R2. In eICU, mean log loss increased by $0.00106$ (0.29426 to 0.29533; 95\% CI for the difference $-0.00087$ to $+0.00355$), corresponding with a fall in AUROC from 0.81920 to 0.81794. With equal-hospital weighting, R3 performed worse than R2 ($+0.00341$, 95\% CI $+0.00039$ to $+0.00644$). The effect varied between hospitals, with R3 outperforming R2 in 57.0\% of the 79 larger hospitals and 48.0\% of all hospitals.

\input{figures/information_increments}

R3 outperformed R2 in 11 of the 30 MetaVision bootstrap resamples, with a median log loss difference of $+0.00114$ (10\textsuperscript{th}--90\textsuperscript{th} percentile $-0.00107$ to $+0.00405$). When source refitting and target-hospital resampling were combined, the patient-weighted difference ranged from $-0.00292$ to $+0.00807$. In contrast, the R2--R1 difference remained negative throughout ($-0.01905$ to $-0.01026$). Expanding the training cohort to include MIMIC-IV likewise failed to produce a clear R3 benefit in eICU.

When trained on the eICU cohort, R3 outperformed R2 both in held-out eICU hospitals and following transfer to MetaVision ($-0.00541$) and MIMIC-IV ($-0.00381$). This advantage remained when the training population was reduced to 7,762 patients to match the size of the MetaVision cohort, with R3--R2 differences of $-0.00844$ in MetaVision and $-0.00202$ in MIMIC-IV.

\subsection{Models incorporating observation patterns were more sensitive to measurement timing}

The distribution of observations across the 48-hour window affected model performance independently of the number retained (Figure~\ref{fig:timing-decomposition}). Measurements spread throughout the window outperformed random retention in all four representations when evaluated in eICU, with differences in log loss ranging from $-0.00681$ for R1 to $-0.04324$ for R3. Retaining the most recent observations improved all representations except R2 under patient weighting. In contrast, removing the final 12 hours was harmful in every tier compared with matched random removal ($+0.00708$ for R2 to $+0.03170$ for R3).

\input{figures/timing_decomposition}

R3 showed greater sensitivity to measurement timing than R2. Relative to R2, recent and spread retention produced larger reductions in log loss for R3 ($-0.04875$ and $-0.03604$), while removal of the final 12 hours resulted in greater deterioration ($+0.02401$). The MIMIC analyses displayed a similar pattern, with R3 and R4 generally exhibiting the greatest sensitivity to measurement timing.

\subsection{Observation patterns retained predictive value in local eICU models}

When models were trained and evaluated within eICU using hospital-grouped cross-validation, R3 outperformed R2, reducing log loss by $0.00402$ (95\% CI $-0.00517$ to $-0.00283$). Equal-hospital weighting yielded similar results. As observation availability was reduced, this advantage diminished. R3 showed no clear benefit at 50\% retention ($-0.00067$), and performed worse than R2 at 25\% retention ($+0.00497$) and after removal of the final 12 hours ($+0.01795$; Figure~\ref{fig:availability}A).

\subsection{Adaptation partially restored the predictive value of observation patterns}

Retraining models with observation availability matched to the evaluation data restored the advantage of R3 over R2 at both 50\% and 25\% retention. R3--R2 log loss differences improved to $-0.00375$ and $-0.00413$, respectively, with both 95\% confidence intervals excluding zero. However, this came at the expense of performance on complete records, where log loss increased by $0.00354$ for R3 retrained at 50\% retention and $0.01753$ at 25\%.

Recalibration using records with observation availability matched to the evaluation data also recovered the R3 advantage at 50\% retention. With slope-and-intercept recalibration, R3 achieved a $0.00228$ lower log loss than R2 (95\% CI for the difference $-0.00340$ to $-0.00104$). At 25\% retention, the difference was smaller and inconclusive ($-0.00021$, 95\% CI $-0.00101$ to $+0.00061$). When the final 12 hours of observations were removed, recalibration narrowed the performance gap. However, R3 still performed worse than R2, returning a log loss difference of $+0.00778$ (95\% CI $0.00564$ to $0.00991$; Figure~\ref{fig:availability}B).

Combining R2 and R3 predictions using the 10,909 eICU calibration patients achieved lower log loss than recalibrating R3 alone, with improvements of 0.00301 on complete records, 0.00638 at 50\% retention, and 0.00929 at 25\% retention. Allowing separate, unconstrained coefficients for R2 and R3 reduced log loss further, by 0.00153, 0.00131, and 0.00113, respectively. As fewer observations were retained, the fitted model assigned progressively less weight to R3 relative to R2, with the R3 coefficient decreasing from 0.498 on complete records to 0.257 at 50\% retention and 0.091 at 25\% retention.

\input{figures/availability_and_calibration}

%% file: tables/external_performance.tex
\begin{table}[H]
\tableformat
\caption{External performance of MetaVision-trained models in MIMIC-IV (n = 21,682) and eICU (n = 43,878). Results average performance across 20 fitting trials using uncalibrated predictions, with negative paired log-loss differences favouring the richer representation. Unadjusted 95\% CIs use patient bootstrapping for MIMIC-IV and hospital bootstrapping for eICU.}
\label{tab:external}

\begin{tabular}{@{}llccccc@{}}
\toprule
Cohort & Model & Log loss & AUROC & Brier & Intercept & Slope \\
\midrule
\multirow{3}{*}{MIMIC-IV}
 & R1 & 0.29787 & 0.84864 & 0.09038 & 0.266 & 1.178 \\
 & R2 & 0.28432 & 0.87039 & 0.08689 & 0.539 & 1.276 \\
 & R3 & 0.27867 & 0.87586 & 0.08538 & 0.467 & 1.250 \\
\addlinespace
\multirow{3}{*}{eICU}
 & R1 & 0.30872 & 0.79049 & 0.09130 & -0.080 & 1.022 \\
 & R2 & 0.29426 & 0.81920 & 0.08680 & 0.073 & 1.149 \\
 & R3 & 0.29533 & 0.81794 & 0.08756 & 0.018 & 1.106 \\
\bottomrule
\end{tabular}

\par\medskip\par

\begin{tabular}{@{}lllcc@{}}
\toprule
Cohort & Difference & Weighting & Estimate & 95\% CI \\
\midrule
\multirow{2}{*}{MIMIC-IV}
 & R2--R1 & Patient
 & $-0.01355$ & \cirange{$-0.01551$}{$-0.01153$} \\
 & R3--R2 & Patient
 & $-0.00564$ & \cirange{$-0.00709$}{$-0.00415$} \\
\addlinespace
\multirow{4}{*}{eICU}
 & R2--R1 & Patient
 & $-0.01446$ & \cirange{$-0.01701$}{$-0.01202$} \\
 & R2--R1 & Equal hospital
 & $-0.01392$ & \cirange{$-0.01782$}{$-0.01003$} \\
 & R3--R2 & Patient
 & $+0.00106$ & \cirange{$-0.00087$}{$+0.00355$} \\
 & R3--R2 & Equal hospital
 & $+0.00341$ & \cirange{$+0.00039$}{$+0.00644$} \\
\bottomrule
\end{tabular}

\end{table}

%% file: figures/information_increments.tex
\begin{figure}[H]
\vspace{22pt}
\centering
\includegraphics[width=\linewidth]{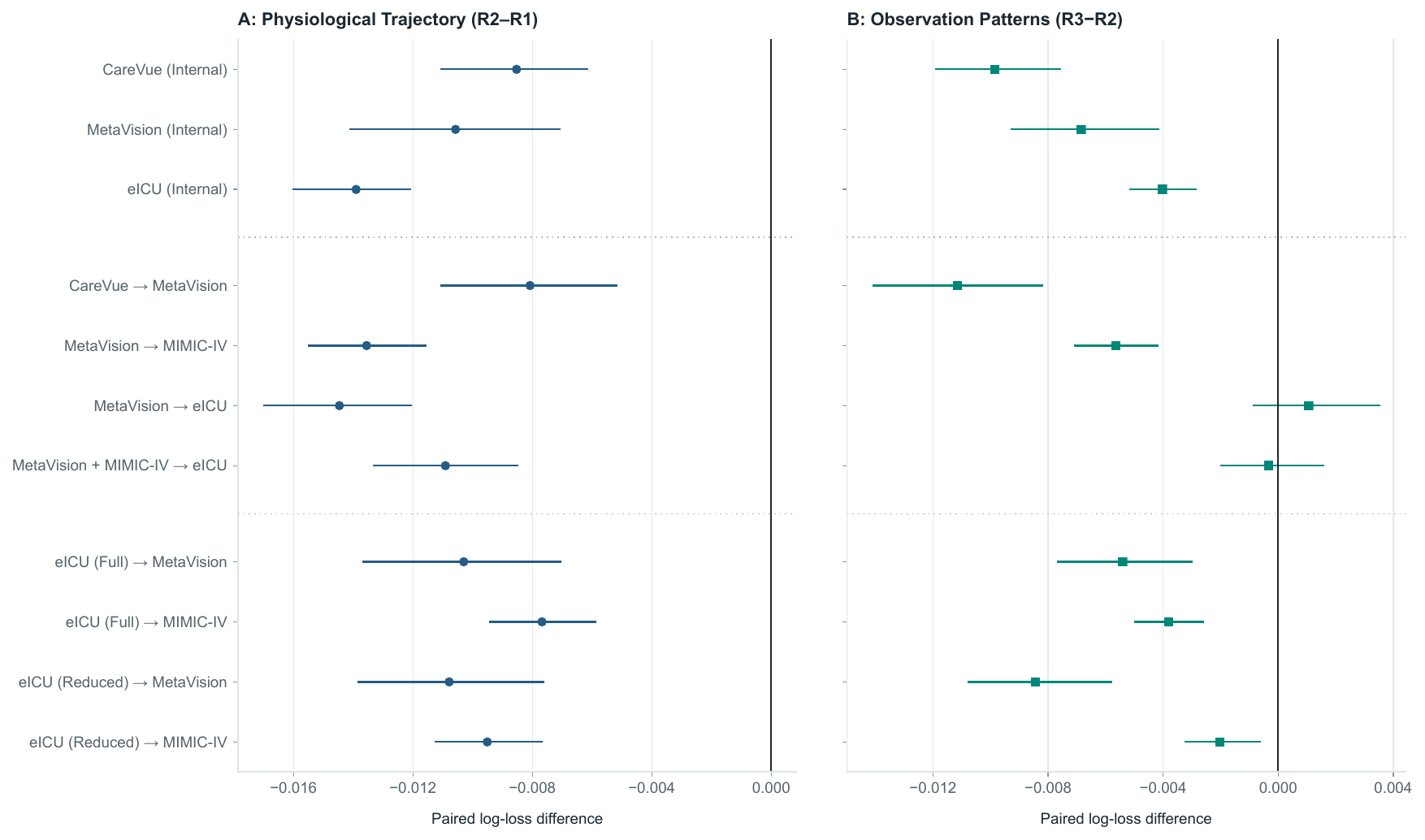}
\caption{
Incremental value of physiological trajectories (R2--R1; A) and observation patterns (R3--R2; B) across evaluation settings. Points show patient-weighted log-loss differences averaged over fitting seeds 0--19, with 95\% paired-bootstrap CIs using patient resampling for MIMIC and hospital resampling for eICU. eICU internal and reduced-source estimates are also averaged over 20 allocations. Negative differences favour the richer representation.
}
\label{fig:increments}
\end{figure}

%% file: figures/timing_decomposition.tex
\begin{figure}[H]
\centering
\includegraphics[width=\linewidth]{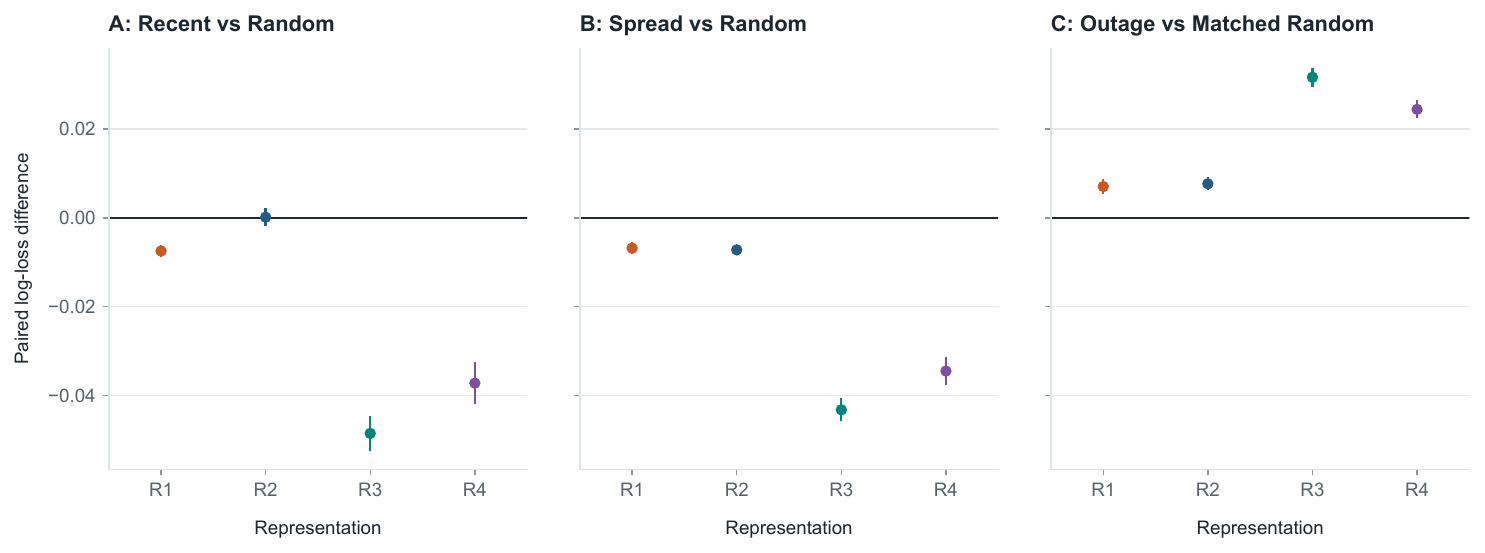}
\caption{External eICU effects of observation timing at matched observation counts. A: most recent observations versus random retention. B: observations spread through the window versus random retention. C: final 12-hour outage versus matched random removal. Negative differences favour recent or spread retention, while positive outage differences indicate worse performance. Points show patient-weighted log-loss differences averaged across restriction iterations, with 95\% hospital-bootstrap CIs.}
\label{fig:timing-decomposition}
\end{figure}

%% file: figures/availability_and_calibration.tex
\begin{figure}[H]
\centering
\includegraphics[width=\linewidth]{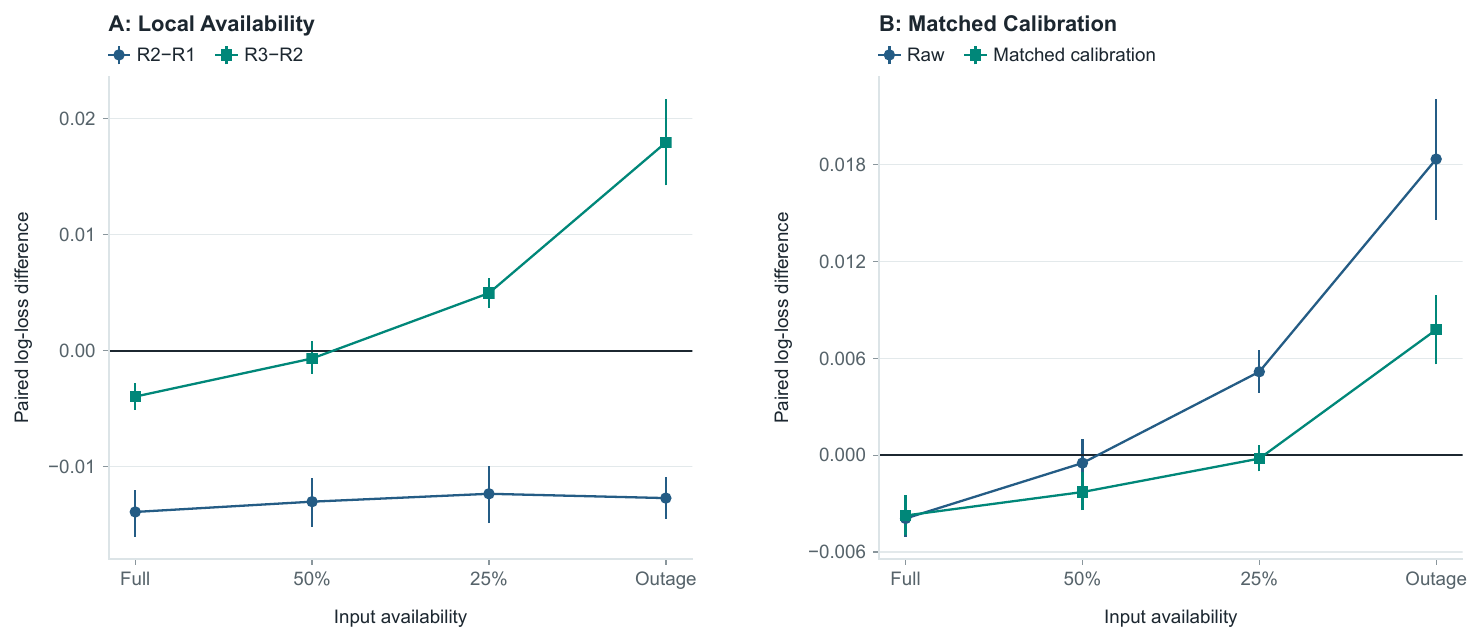}
\caption{
Local eICU effects of changing observation availability. A: R2--R1 and R3--R2 differences when models trained on complete records are evaluated on complete or restricted records. B: R3--R2 differences before and after restriction-matched recalibration using hospitals reserved for calibration within each training fold. Points show patient-weighted log-loss differences averaged across allocations, with 95\% hospital-bootstrap CIs.
}
\label{fig:availability}
\end{figure}

%% file: sections/discussion.tex
\section{Discussion}

\subsection{Interpretation of findings}

Longitudinal physiological information improved ICU mortality prediction consistently across recording systems, periods, and institutions. The additional value of observation patterns proved less reliable. When trained on single-institution cohorts, models demonstrated no clear benefit from observation patterns when transferred to multi-institution cohorts, whereas benefits were retained internally and in the reverse direction. The distribution of measurements within the observation window influenced performance independently of their count. Representations incorporating observation patterns were particularly sensitive to this phenomenon. These models were similarly sensitive to changes in observation availability, although performance could be partially recovered through adaptation.

The asymmetric transfer results suggest that training across multiple institutions may allow models to capture the informative value of observation patterns without relying on associations specific to individual clinical settings. Models trained on eICU retained the benefit of these features in held-out hospitals and following transfer to MetaVision and MIMIC-IV, even when the training population was reduced to match MetaVision in size. Conversely, expanding the MetaVision cohort to include MIMIC-IV did not produce a clear benefit in eICU. Together, these findings indicate that the diversity in measurement practices represented in the training data may contribute more to generalisation than cohort size alone.

Previous studies have reported that observation patterns carry prognostic information \citep{agniel2018,zhang2019} while also highlighting their sensitivity to changes in clinical documentation and measurement practices \citep{nestor2019,groenwold2020}. Recent findings have raised similar concerns about their transferability \citep{yamamoto2026,zou2026}. However, these limitations may not be inherent to the features themselves and could be mitigated through approaches to model development and adaptation that better account for variation in observation practices.

However, reliance on observation patterns may introduce sensitivity not only to how frequently measurements are recorded, but also to when they occur. At matched observation counts, retaining recent or evenly distributed measurements produced substantial improvements over random retention for representations incorporating observation patterns, whereas the simpler representations were less sensitive. Removing recent measurements similarly caused far greater deterioration in these models. These results suggest a particular dependence on the recency, spacing, and coverage of measurements, rendering their performance vulnerable to changes in the temporal structure of clinical records. This highlights a potential cost of incorporating observation patterns, consistent with broader concerns about sensitivity to changing missingness patterns \citep{rockenschaub2024,zhou2023}.

Changes in measurement availability further highlighted the sensitivity of observation patterns to differences between the development and evaluation conditions. As availability decreased, performance declined in models incorporating observation patterns, relative to those using longitudinal information alone, and their predictions were progressively downweighted when the two were combined. However, retraining on similarly restricted records restored its advantage, suggesting that the deterioration stemmed from a mismatch in measurement conditions rather than reduced availability alone. Whilst recalibration also recovered some of the benefit, retraining compromised performance on complete records. Existing research has considered model updating and recalibration in clinical prediction \citep{steyerberg2004,vancalster2019}, including adaptation to differences between clinical environments \citep{kappen2012,davis2019}. Nonetheless, the limited recovery seen under severe restrictions underscores the challenges that remain in adapting models to more substantial changes in measurement conditions. Similar difficulties have been reported in studies of domain adaptation under temporal dataset shift \citep{guo2022}.

\subsection{Strengths and limitations}

Several aspects of the study design help establish the reliability of these findings. Comparisons across recording systems, time periods, and institutions tested whether the benefits of richer representations extended beyond the settings in which models were developed. Local eICU development and reverse transfer helped distinguish whether observation patterns lacked predictive value or whether that value simply failed to generalise across settings. By matching training cohort sizes, the analysis also assessed whether differences in sample size alone could explain the contrasting transfer results. Likewise, the restriction experiments helped separate the effects of measurement timing from those of observation count. Finally, repetition of restriction trials, model fitting, and hospital-grouped evaluations reduced the dependence on individual random selections, while source refitting and hospital-level resampling supported the stability of the external results.

There are, however, some important limitations to consider. Despite harmonisation, MIMIC and eICU differ in patient populations, clinical practices, and variable definitions. It is therefore not possible to separate the influence of institutional diversity from that of other cohort characteristics. Similarly, comparisons between representations do not isolate effects of individual features, since their inclusion also affects model interactions and regularisation. Although the timing experiments controlled observation counts, selecting different measurements inevitably changed the physiological information available to the models. Furthermore, these restrictions were simulated retrospectively and may not reflect actual monitoring practices, as decisions about when to measure patients are influenced by their clinical condition. Finally, the analysis was restricted to patients remaining in intensive care at 48 hours, limiting its applicability to other populations and prediction times.

\subsection{Implications and future directions}

Models may generalise across patient populations without necessarily generalising across measurement practices. External validation should therefore consider both. An important next step is to establish whether greater institutional diversity during training improves the transferability of information from observation patterns, independently of sample size and patient characteristics. Comparisons across hospitals with documented differences in monitoring practices could also help identify which aspects of measurement frequency and timing are most consequential for prediction. Further work should examine how effectively models can be adapted to new measurement environments, particularly when local data are limited, and whether similar challenges arise with other outcomes and prediction times. Ultimately, understanding how changes in the measurement process affect prediction is essential to establishing when observation patterns can be used with confidence.

%% file: sections/conclusion.tex
\section{Conclusion}

Longitudinal physiological information improved ICU mortality prediction across recording systems, periods, and institutions. Observation patterns provided additional predictive value, however their contribution was less consistent across settings and particularly sensitive to changes in measurement timing and availability. Performance could be partially recovered through adaptation, suggesting that this vulnerability arises in part from differences in measurement conditions between development and evaluation, rather than from an inherent lack of predictive information. Training across multiple institutions may help reduce this dependence, although its contribution requires further investigation. Clinical prediction models should therefore be evaluated not only across patient populations, but also under the measurement conditions in which they are intended to operate.

%% file: sections/declarations.tex
\section*{Declarations}

This study involved secondary analysis of de-identified clinical records from MIMIC-III, MIMIC-IV, and eICU. The MIMIC databases were developed with institutional ethics approval and waivers of informed consent. The eICU database was de-identified in accordance with HIPAA requirements. Data were accessed under the applicable credentialing and data-use agreements.

The datasets analysed in this study are available through PhysioNet: \href{https://physionet.org/content/mimiciii/1.4/}{MIMIC-III version 1.4}, \href{https://physionet.org/content/mimiciv/3.1/}{MIMIC-IV version 3.1}, and \href{https://physionet.org/content/eicu-crd/2.0/}{eICU version 2.0}. Access requires completion of the relevant training and acceptance of the PhysioNet data-use agreements. The analysis code is available on \href{https://github.com/benjamin-pike/icu-mortality-prediction}{GitHub} and archived on Zenodo (\href{https://doi.org/10.5281/zenodo.23741498}{doi:10.5281/zenodo.23741498}). Full results data are available on Zenodo (\href{https://doi.org/10.5281/zenodo.23741579}{doi:10.5281/zenodo.23741579}).

This research received no external funding and the author declares no competing interests. The author was responsible for study conception, methodology, software development, data analysis, interpretation, and manuscript preparation.

During the preparation of this manuscript, OpenAI's ChatGPT was used to assist with grammatical proofreading and refactoring of analysis code originally written by the author. All changes were reviewed by the author, who takes full responsibility for the final manuscript and analyses.